\documentclass{article} 

\pdfoutput=1

\usepackage{graphicx}
\usepackage{amsmath,amssymb}
\usepackage[american]{babel}
\usepackage{float}
\usepackage{subfigure}
\newcommand{\real}{\mathbb{R}}
\def\argmin{\mathop{\rm argmin}}

\title{A mixture Cox-Logistic model for feature selection from survival and classification data}

\author{Samuel Branders, Roberto D'Ambrosio and Pierre Dupont\\\ \\
Universit\'e catholique de Louvain\\
ICTEAM -- Machine Learning Group\\
Place Sainte Barbe 2, B-1348 Louvain-la-Neuve, Belgium.\\
}

\begin{document}
 
\maketitle

\begin{abstract}
This paper presents an original approach for jointly fitting survival times and classifying samples into subgroups.
The Coxlogit model is a generalized linear model with a common set of selected features for both tasks.
Survival times and class labels are here assumed to be conditioned by a common risk score which depends on those features.
Learning is then naturally expressed as maximizing the joint probability of subgroup labels and the ordering of survival events, conditioned to a common weight vector.
The model is estimated by minimizing a regularized log-likelihood  through a coordinate descent algorithm.

Validation on synthetic and breast cancer data shows that the proposed approach outperforms 
a standard Cox model or logistic regression when both predicting the survival times and classifying new samples into subgroups. It is also better at selecting informative features for both tasks.
\end{abstract}

\section{Introduction}

Survival analysis aims at modeling the relationships between several covariates (\textit{e.g.} age, gender, environmental factors, gene expression values, \dots) and the time of specific events, such as relapse, metastasis or death. Cox proportional hazard models are often used towards this objective~\cite{Cox1972}.
A distinct objective is to map the (patient) samples to different subgroups, for instance corresponding to specific tumor grades or subtypes.
Given a collection of such samples labeled by clinicians, this second problem reduces to supervised learning of a classifier for which 
any standard algorithm (SVM, logistic regression, Naive Bayes, \dots) could be used.

The originality of this work is to tackle both problems jointly since the specific subgroups of interest may exhibit distinct risk profiles which, in turn, 
could condition their survival times~\cite{Sotiriou2006}. We consider in particular generalized linear models as they offer a direct interpretation in terms of individual feature relevances. The proposed Coxlogit model is a natural extension to logistic regression for which we assume that the survival times 
and class labels are random variables conditioned by a common risk. We show that the partial likelihood of such model, to fit the ordering of observed survival times, is directly related to the logistic class probabilities. Learning can then be expressed as maximizing the joint probability of class labels and the ordering of survival events, conditioned to a common weight vector. 
Embedded feature selection follows naturally when fitting such a model with a LASSO or elastic net penalty. 
Such penalties prevent overfitting while enforcing a common sparse support.   
Learning is also a convex problem that can be efficiently solved through a coordinate descent algorithm.

We report practical experiments both on synthetic and real breast cancer datasets.
Those experiments show that the Coxlogit approach outperforms either a Cox model or logistic regression
when both predicting the survival times and classifying new samples into subgroups.
The proposed approach is also better at selecting features that are informative for both tasks simultaneously.

\section{The Coxlogit approach}

One considers a survival analysis framework made of a collection of samples and their associated survival times, which are possibly censored.
One further assumes that each training sample is labeled into a specific subgroup.
Formally, each sample $i \in \{1, \dots , n\}$ is characterized by a 4-tuple $(t_i,\delta_i,y_i,x_i)$ where $t_i$ is the time of an event (such as metastasis or relapse), whenever $\delta_i=1$, and the censoring time whenever $\delta_i=0$. Furthermore, $y_i$ denotes a binary class label, respectively $-1$ and $1$ for two subgroups of interest. Patients of class $1$ are expected to have a higher risk than patients of the class $-1$.

The survival data and the class label of patient $i$ are seen here as two observations of random variables conditioned by a common risk of event, $r_i$.
This risk is simply modeled as a linear combination of the sample covariates ($x_i \in \real^p$): $r_i = \beta^\top x_i$
but the fit of the parameters $\beta$ should consider both types of supervision.

Starting from the classification viewpoint, a logistic regression predicts from the vector $x_i$ the probability of patient $i$ to be in a specific group:

\begin{eqnarray}
    P(Y_i=1 | x_i) &=& \frac{\exp(\beta^\top x_i) }{1 + \exp(\beta^\top x_i)} \\
    P(Y_i=-1 | x_i) &=& \frac{1}{1 + \exp(\beta^\top x_i)} = 1 - P(Y_i=1 | x_i)
\end{eqnarray}

The risk score of a patient, $r_i = \beta^\top x_i$, can be interpreted through the logistic model as class probabilities: high risk patients are more likely to be in the high risk group +1 
and a zero risk score corresponds to an equal probability to be in either subgroups.
The likelihood of the parameters $\beta$ with respect to the observed labels $y_i$ is given by:

\begin{eqnarray}
    L(\beta) = \prod_{i=1}^n P(Y_i=y_i | x_i) = \prod_{i=1}^n \frac{ 1 }{1 + \exp(-y_i (\beta^\top x_i))}
\end{eqnarray}

Looking now at the survival times and knowing that an event occurs at $t_i$, one typically computes the probability of patient $i$ having the event over the set of patients still at risk just before time $t_i$ : $R(t_i)= \{j | t_j \geq t_i\}$.
Since high risk patients tend to have the event before low risk patients, the likelihood of observed events can also be seen as the conditional probability of patient $i$ being in the high risk group and all others in the low risk group (knowing that exactly one patient has the event before the others at that time).
This likelihood can also be expressed in terms of the logistic class probabilities $P(Y_i=1 | x_i)$ and $P(Y_i=-1 | x_i)$ :

\begin{eqnarray}
    L_i(\beta) &=& \frac{ P(Y_i=1 |x_i) \prod_{j \in R(t_i)\backslash \{i\}} P(Y_j = -1 | x_j) }{ \sum_{k \in R(t_i)} P(Y_k=1 |x_k) \prod_{j \in R(t_i)\backslash \{k\}} P(Y_j = -1 | x_j) } \label{condprob} \\
    &=& \frac{ \frac{\exp(\beta^\top x_i) }{1 + \exp(\beta^\top x_i)} \prod_{j \in R(t_i)\backslash \{i\}} \frac{1}{1 + \exp(\beta^\top x_j)} }{ \sum_{k \in R(t_i)} \frac{\exp(\beta^\top x_k) }{1 + \exp(\beta^\top x_k)} \prod_{j \in R(t_i)\backslash \{k\}} \frac{1}{1 + \exp(\beta^\top x_j)} }  \\
    &=& \frac{ \exp{(\beta^\top x_i)}  }{ \sum_{k \in R(t_i)} \exp{(\beta^\top x_k)} } \label{condprob:2}
\end{eqnarray}

Expression~(\ref{condprob:2}), aggregated over all survival times, boils down to the partial\footnote{It is called \textsl{partial} as it only relies on the ordering of the events and not the actual times when those events occur.} likelihood of a Cox model for survival data,
except that censoring should also be considered. Formally, the computation is restricted to those patients for which the event is observed ($\delta_i = 1$).

The likelihood of the Coxlogit model is now defined as the joint probability of the observed events and subgroup labels knowing the parameters $\beta$.
Assuming the labels and the times to event to be conditionally independent \textsl{given} those parameters, this likelihood can be computed as:

\begin{eqnarray}
L(\beta) = \prod_{i=1}^n \frac{ 1 }{1 + \exp(-y_i (\beta^\top x_i))} \left[ \frac{ \exp{(\beta^\top x_i)}  }{ \sum_{j \in R(t_i)} \exp{(\beta^\top x_j)} } \right]^{\delta_i}
\end{eqnarray}

The log-likelihood $l(\beta)$ of the Coxlogit model is thus naturally formulated as a mixture of a Cox and logistic regression log-likelihoods:

\begin{eqnarray}
- l(\beta) &=& \sum_{i=1}^n \log(1 + \exp(-y_i \beta^\top x_i)) \nonumber\\
&&- \sum_{i=1}^n \delta_i \beta^\top x_i + \sum_{i=1}^n \delta_i \log( \sum_{j \in R(t_i)} \exp{(\beta^\top x_j)} ) 
\label{objective}
\end{eqnarray}

An embedded feature selection follows by regularizing this objective:
\begin{equation}
\argmin_{\beta} -\frac{2}{n} l(\beta) + \lambda \; \Omega(\beta)
\label{reg:objective}
\end{equation}
\noindent
where $\Omega(\beta)$ is a sparsity enforcing regularization such as LASSO or elastic net~\cite{Zou2005}, and $\lambda > 0$ a regularization constant. A coordinate descent algorithm, adapted from~\cite{Simon2011}, is used here to solve this convex problem. It starts from a trivial solution ($\beta=0$) for a large $\lambda$, and follows the regularization path when $\lambda$ is gradually decreased till the model includes a desired number of features (= non-zero weight values).

\section{Experiments\label{sec:Experiments}}

We first consider an artificial dataset to assess to which extent the Coxlogit approach 
is able to select informative features both for classification and survival prediction.
A data matrix $X \in \real^{n\times p}$ is drawn from a $\mathcal{N}(0,1)$ distribution to represent
covariates that have been centered and normalized to unit variance. 
Those features are partitioned into 4 groups. Each of the 3 first groups includes $k$ features 
which are predictive either of the survival, the group label or both. The $p-3k$ remaining features 
are purely random and represent noise.

The hazards\footnote{In survival analysis, the hazard is a time dependent function corresponding to the probability of a patient, still at risk, to experience the event at time $t$.} and group assignments are generated from distinct linear combinations of the informative features,
which are drawn from a uniform distribution over $[-1,-0.5] \cup [0.5,1]$.
The survival data $(t_i, \delta_i)$ are generated from two weibull distributions. The distribution for the time to event $t_i$ 
is parametrized such that the hazards $h_i(t)$ only depend on the features from the first and second group: $h_i(t) \propto \exp(X\beta_{1\dots k\; (k+1) \dots 2k}^{\top})$.
Similarly, the group labels $y \in \{-1,1\}^n$ only depend on the features from the first and third groups: $y = \mathrm{sign} (X \beta_{1\dots k\, (2k+1) \dots 3k}^{\top})$.

Results are averaged in the table below over 100 independent runs with $n=1000$ samples (200 for training, 800 for validation) with $p=100$ features among which the first $k=10$ features are jointly predictive of survival and classification. The Coxlogit model is compared to a Cox proportional hazard model and logistic regression while following in all cases the regularization path till selecting exactly $10$ features.
The first column illustrates that the Coxlogit approach selects more features from the first group,
which are those informative for group classification and survival prediction.
The remaining columns report the validation results in terms of classification accuracy, Concordance 
index (measuring the correct ordering of survival times according to risk groups~\cite{Harrell1996}) and the harmonic average between those metrics. They illustrate that Coxlogit outperforms the other approaches when tackling both tasks.

\begin{center}
\begin{tabular}{|c||c|c|c|c|}
  \hline
  Methods & Features & Accuracy & C-index & Predictive Performance \\
  \hline
  Coxlogit & \textbf{6.59/10} & 0.67 & 0.80 & \textbf{.73}\\
  Cox & 4.67/10 & 0.59 & 0.81 & .68 \\
  Logistic & 4.65/10 & 0.71 & 0.65 & .68 \\
  \hline
\end{tabular}
\end{center}

We further assess the Coxlogit approach on 4 breast cancer studies (GSE2034, GSE5327, GSE7390, GSE2990) from the GEO repository. Those samples are gene expression values measured on the Affymetrix HGU133a microarray platform and distant metastasis is used as survival end point. All samples are gathered in a common dataset including $554$ patients and $1236$ features, after keeping only the dimensions with the largest variances. The objective is to predict both the grade of the tumor~\cite{Galea1992}, discretized into low versus high grade with roughly equal priors, and the survival probability of the patients.

Results are reported below over 100 resamplings (without replacement) into 90\% training/10\% test
over various feature set sizes. The predictive performance (averaged over 100 runs) is 
the harmonic mean between test classification accuracy and Concordance index. Such an aggregate metric is 
representative of the performances on both tasks of grade classification and survival prediction.
Those results illustrate the overall benefit of the Coxlogit model as compared to the original 
Cox model or logistic regression.   

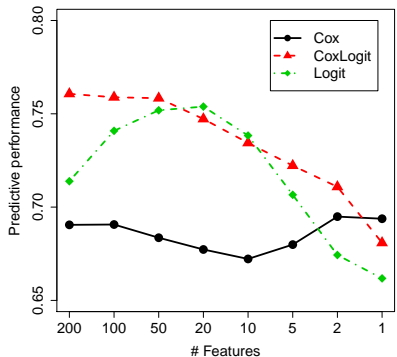
\vspace*{-4mm}   
\centerline{\includegraphics[width=.45\textwidth,page=8]{RplotsSurvivalSimulation8.pdf}}
\vspace*{-4mm}

\section{Conclusion and perspectives}

This paper describes the Coxlogit method which is a generalized linear model to predict survival times 
and jointly classify samples into subgroups. Once regularized with a sparsity inducing term,
it offers an embedded feature selection to discover informative features for both tasks. We consider here classification into 2 specific groups but generalization to multi-class or continuous response looks interesting and easy.
It would essentially amount to replace the logistic part of the objective by its multinomial extension using a softmax function or by a square loss. 
The specific subgroups of interest are here supposed \textsl{a priori} known at training time, and typically provided by clinical annotations in a real scenario.
Such assumption could also be relaxed by considering unsupervised or semi-supervised learning of those groups.

\section*{Acknowledgments}

\paragraph{Funding} The work of Samuel Branders is supported by the F.R.S. - FNRS - T\'el\'evie (Grant number FC 88088).

\bibliographystyle{abbrv}
\bibliography{library}

\end{document}